%% file: 00_main.tex
\documentclass[10pt,journal]{IEEEtran}
\ifCLASSOPTIONcompsoc
  \usepackage[nocompress]{cite}
\else
  \usepackage{cite}
\fi
\ifCLASSINFOpdf
\else
\fi
\usepackage{amsmath}

\usepackage{todonotes}

\usepackage{subcaption}
\usepackage{colortbl}
\usepackage{multirow}

\usepackage[normalem]{ulem}

\usepackage{tikz}
\usepackage{textcomp}
\usepackage{hyperref}
\usepackage{lipsum}

\newcommand\copyrighttext{%
  \footnotesize \textcopyright 2017 IEEE. Personal use of this material is permitted. Permission from IEEE must be obtained for all other uses, in any current or future media, including reprinting/republishing this material for advertising or promotional purposes, creating new collective works, for resale or redistribution to servers or lists, or reuse of any copyrighted component of this work in other works. DOI: \href{https://dx.doi.org/10.1109/JBHI.2017.2785682}{10.1109/JBHI.2017.2785682}}
\newcommand\copyrightnotice{%
\begin{tikzpicture}[remember picture,overlay]
\node[anchor=south,yshift=5pt] at (current page.south) {\fbox{\parbox{\dimexpr\textwidth-\fboxsep-\fboxrule\relax}{\copyrighttext}}};
\end{tikzpicture}%
}

\begin{document}

%
\title{Automatic Classification of Functional Gait Disorders}

%
%
%
%

\author{Djordje Slijepcevic, Matthias Zeppelzauer, Anna-Maria Gorgas, Caterine Schwab, Michael Sch{\"u}ller, Arnold Baca, Christian Breiteneder, Brian Horsak

\thanks{Manuscript received --; revised --.}

\thanks{Djordje Slijepcevic and Matthias Zeppelzauer are with St. P\"olten University of Applied Sciences, Austria and TU Wien, Austria.}

\thanks{Anna-Maria Gorgas, Caterine Schwab, and Brian Horsak are with St. P\"olten University of Applied Sciences, Austria.}

\thanks{Michael Sch{\"u}ller and Arnold Baca are with University of Vienna, Austria.}

\thanks{Christian Breitender is with TU Wien, Austria.}

}

%
%

\markboth{IEEE Journal of Biomedical and Health Informatics, December~2017}%
{Slijepcevic \MakeLowercase{\textit{et al.}}: Automatic Classification of Functional Gait Disorders}
%



\IEEEtitleabstractindextext{%
\begin{abstract}
This article proposes a comprehensive investigation of the automatic classification of functional gait disorders based solely on ground reaction force (GRF) measurements. The aim of the study is twofold: (1) to investigate the suitability of state-of-the-art GRF parameterization techniques (representations) for the discrimination of functional gait disorders; and (2) to provide a first performance baseline for the automated classification of functional gait disorders for a large-scale dataset. 
The utilized database comprises GRF measurements from 279 patients with gait disorders (GDs) and data from 161 healthy controls (N). 
Patients were manually classified into four classes with different functional impairments associated with the ``hip", ``knee", ``ankle", and ``calcaneus".
Different 
parameterizations are investigated: GRF parameters, global principal component analysis (PCA)-based representations and a combined representation applying PCA on GRF parameters. The discriminative power of each parameterization for different classes is investigated by linear discriminant analysis (LDA).
Based on this analysis, two classification experiments are pursued: 
(1) distinction between healthy and impaired gait (N vs. GD) and (2) multi-class classification between healthy gait and all four GD classes. 
Experiments show promising results and reveal among others that several factors, such as imbalanced class cardinalities and varying numbers of measurement sessions per patient have a strong impact on the classification accuracy and therefore need to be taken into account. 
The results represent a promising first step towards the automated classification of gait disorders and a first performance baseline for future developments in this direction. 







\end{abstract}

}

\maketitle

\IEEEdisplaynontitleabstractindextext

%
\IEEEpeerreviewmaketitle

\copyrightnotice

\section{Introduction}
\label{sec:introduction}
\input{01_intro}

\section{Material and Methods}
\label{sec:methods}
\input{04_methods}

\section{Results and Discussion}
\label{sec:results}
\input{05_results}

\section{Conclusions}
\label{sec:conclusions}
\input{06_conclusion}

\ifCLASSOPTIONcompsoc
  \section*{Acknowledgments}
This project is funded by the NFB – Lower Austrian Research and Education Company and the Provincial Government of Lower Austria, Department of Science and Research (LSC14-005). We want to thank Marianne Worisch and Szava Zolt\'{a}n for their great assistance in data preparation and their great support in clinical and technical questions.
\else
  \section*{Acknowledgment}
  This project is funded by the NFB – Lower Austrian Research and Education Company and the Provincial Government of Lower Austria, Department of Science and Research (LSC14-005). We want to thank Marianne Worisch and Szava Zolt\'{a}n for their great assistance in data preparation and their great support in clinical and technical questions.
\fi
\ifCLASSOPTIONcaptionsoff
  \newpage
\fi
%
\bibliographystyle{unsrt}
\bibliography{07_refs}

\setcounter{figure}{0}  

\begin{figure*}[!t]
\centering
\begin{subfigure}[$F_{V}$]{\textwidth}\centering
	\includegraphics[width=0.54\textwidth]{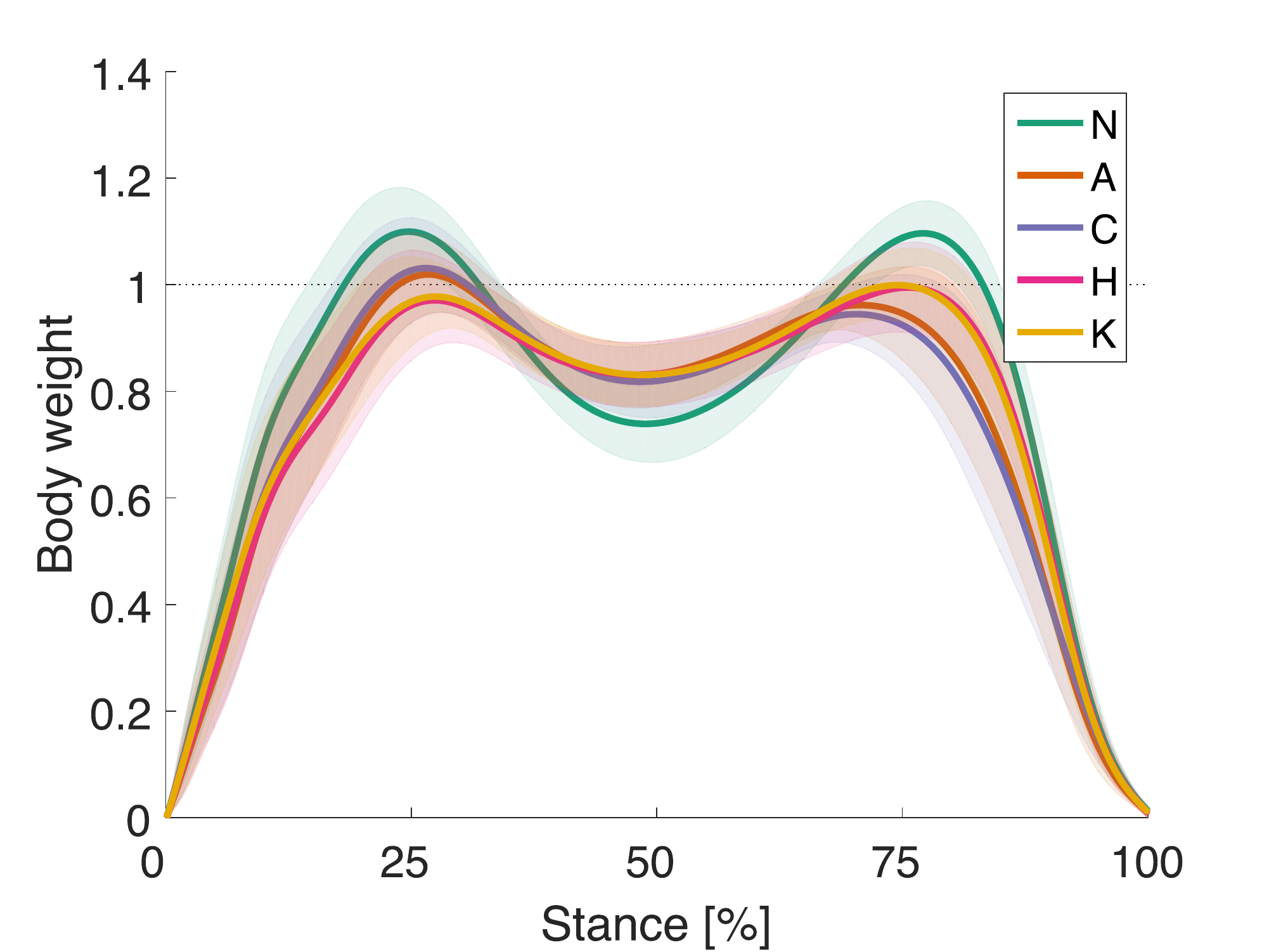}
    \caption{$F_{V}$}
    \end{subfigure}
\hfil
\begin{subfigure}[$F_{AP}$]{\textwidth}\centering
	\includegraphics[width=0.54\textwidth]{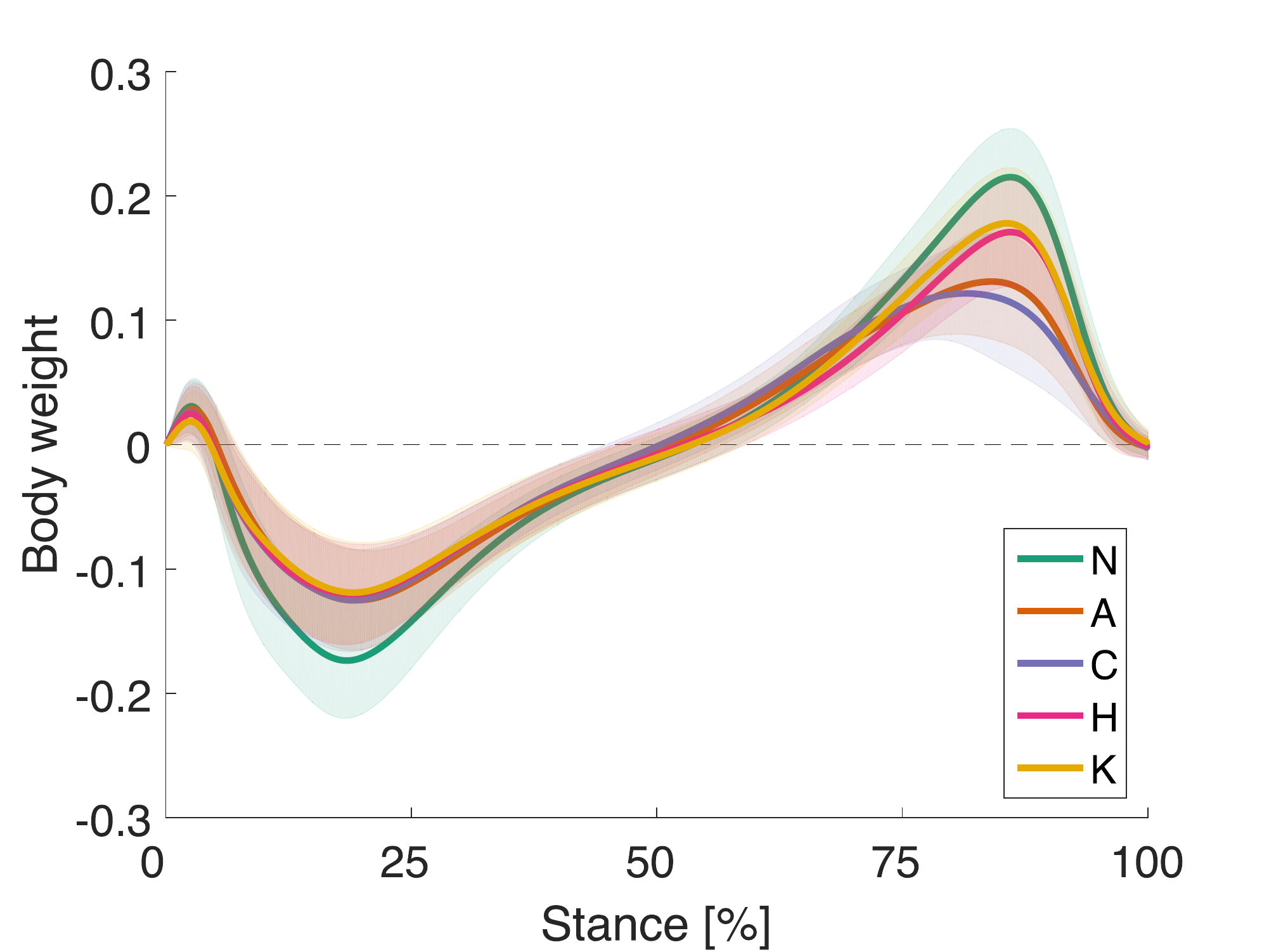}
    \caption{$F_{AP}$}
    \end{subfigure}
\hfill
\begin{subfigure}[$F_{ML}$]{\textwidth}\centering
	\includegraphics[width=0.54\textwidth]{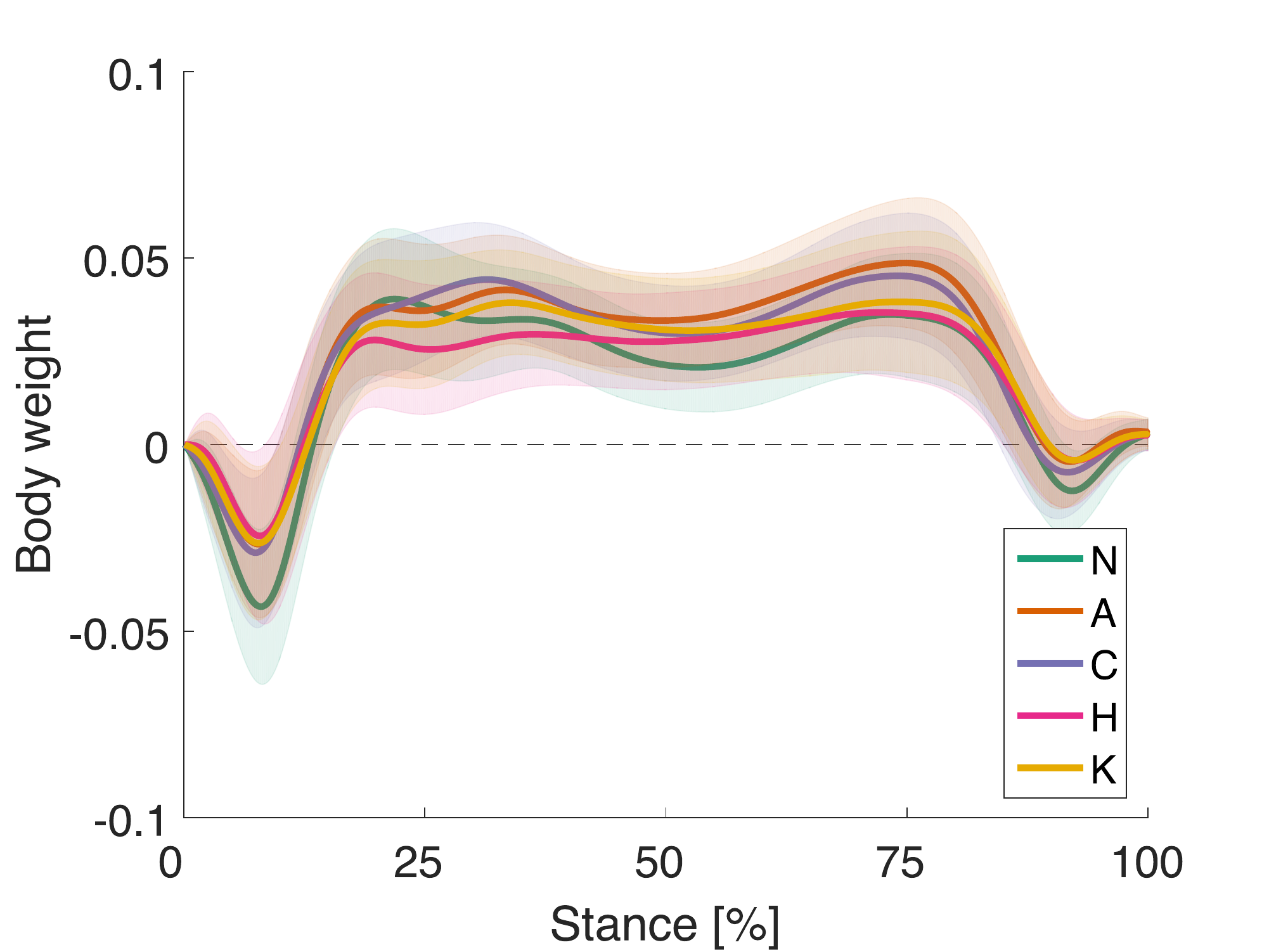}
    \caption{$F_{ML}$}
    \end{subfigure}
\caption{Mean pattern of the three ground reaction forces (GRF) enveloped by $\pm$~1 standard deviations for each class. Data normalized by body weight and 100\% stance.}
\end{figure*}

\begin{figure*}[!t]
\centering
\begin{subfigure}[$F_{V}$]{\textwidth}\centering
	\includegraphics[width=0.54\textwidth]{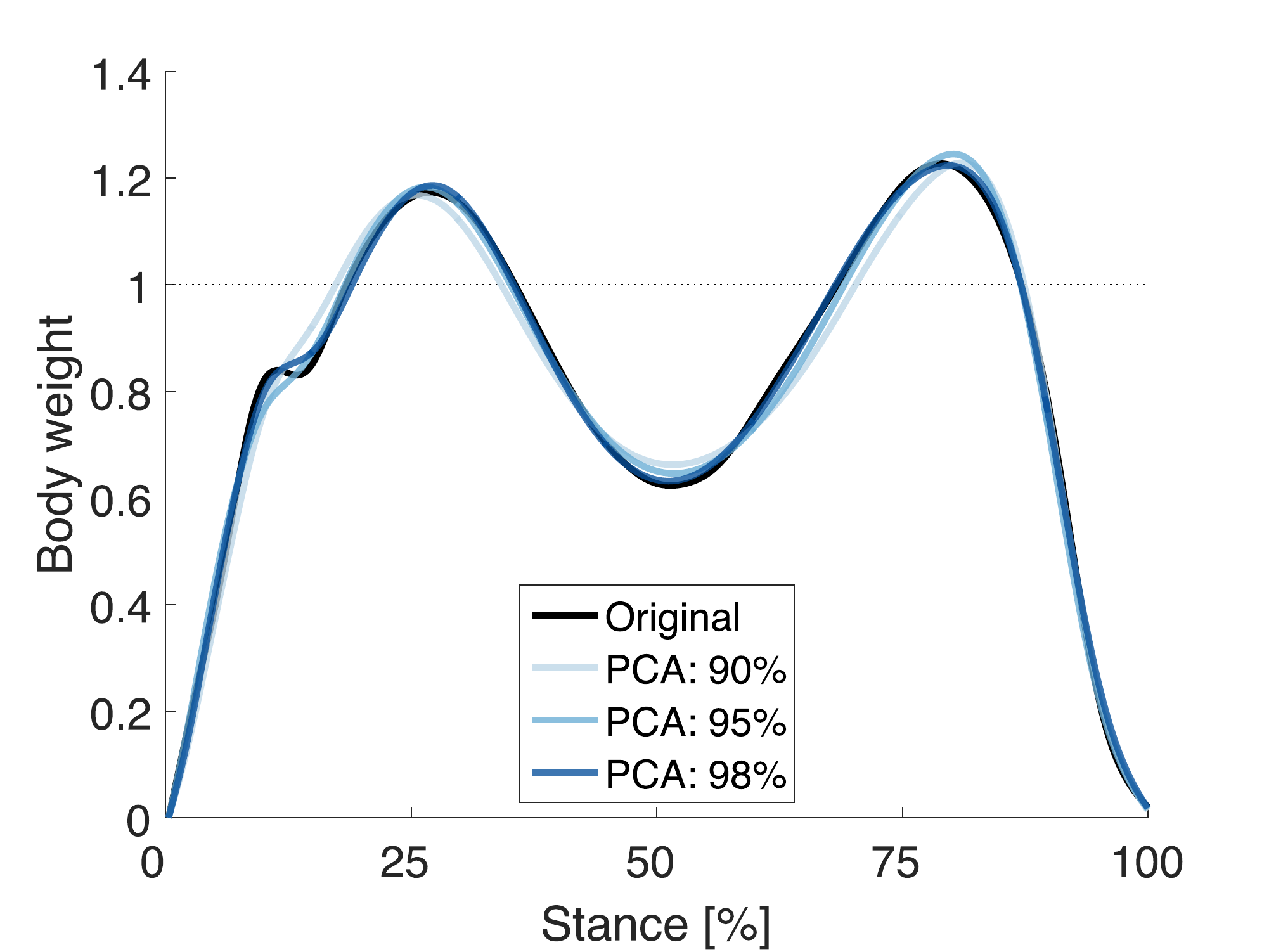}
    \caption{$F_{V}$}
    \end{subfigure}
\hfil
\begin{subfigure}[$F_{AP}$]{\textwidth}\centering
	\includegraphics[width=0.54\textwidth]{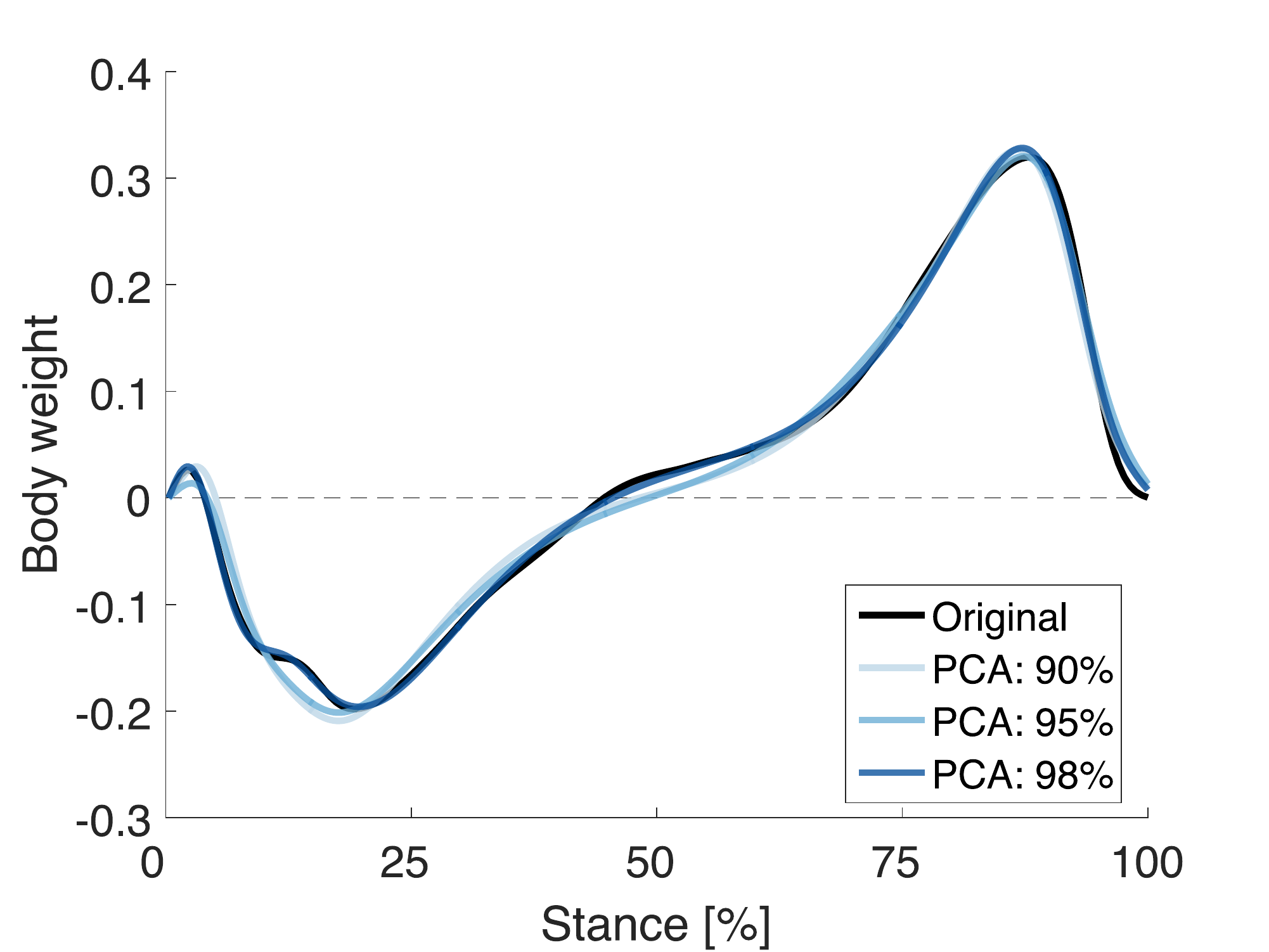}
    \caption{$F_{AP}$}
    \end{subfigure}
\hfill
\begin{subfigure}[$F_{ML}$]{\textwidth}\centering
	\includegraphics[width=0.54\textwidth]{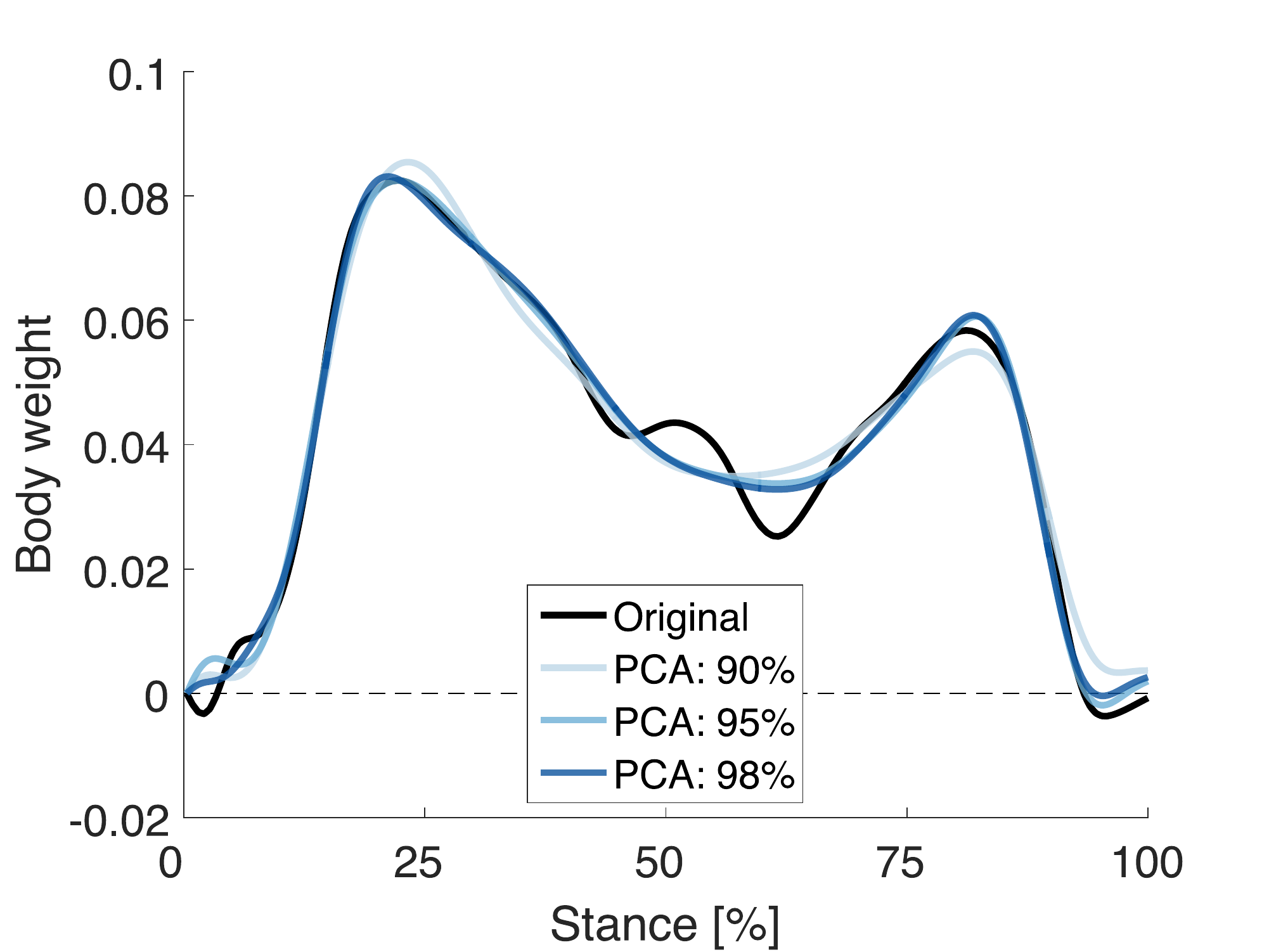}
    \caption{$F_{ML}$}
    \end{subfigure}
\caption{Comparison of different PCA representations. The final dimensionality of the obtained representations is specified by the amount of variance preserved in a particular projection, i.e. 98\%, 95\%, and 90\%. Data normalized by body weight and 100\% stance.}
\end{figure*}

\begin{figure*}[ht!]
  \centering
	\includegraphics[width=\textwidth, height=\dimexpr\textheight-3\baselineskip-\abovecaptionskip-\belowcaptionskip\relax, keepaspectratio]{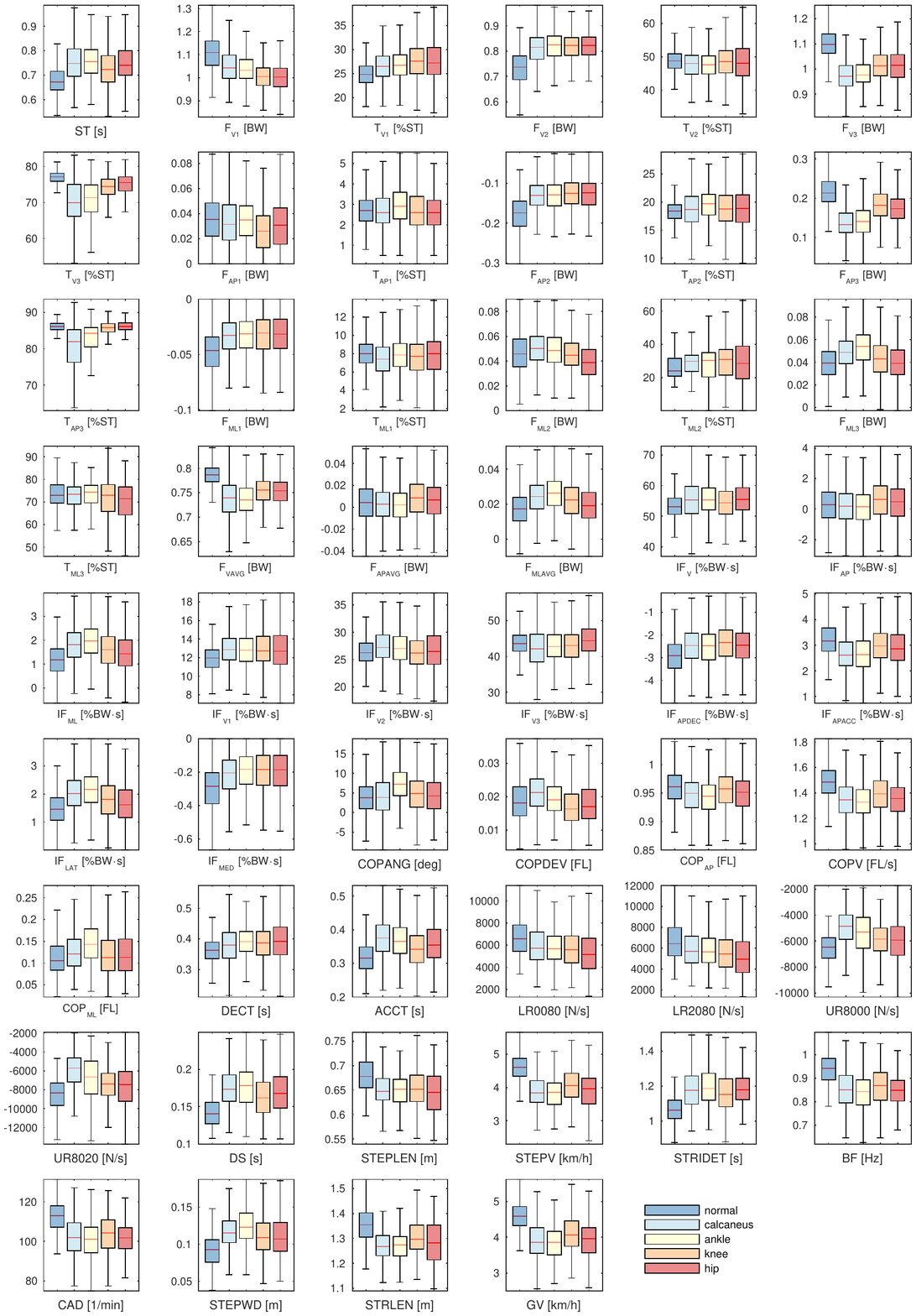}
    \caption{Boxplots for all 52 GRF parameters. Each boxplot shows the median and the IQR (box) for each class (outliers were removed for better visualization). Box-whiskers correspond to 1.5 of the box-length, thus show approximately $\pm$~2.7 standard deviations. The overlap of distributions between the classes gives an impression of the parameters' discriminative power (inter-class variation). Data normalized by body weight and 100\% stance, prior to the calculation of the parameters.}
\end{figure*}

\end{document}

%% file: 01_intro.tex
\IEEEPARstart{G}{ait} analysis is a tool for clinicians to objectively quantify human locomotion and to describe and analyze a patient`s gait performance. The primary aim is to identify impairments that affect a patient's gait pattern~\cite{baker_measuring_2013}. 

Recordings obtained during clinical gait analyses produce a vast amount of data which are difficult to comprehend and analyze due to their high-dimensionality, temporal dependences, strong variability, non-linear relationships and correlations within the data\cite{chau_review_2001-1}. This makes data interpretation challenging and requires an experienced clinician to draw valid conclusions. Several automatic analysis approaches based on machine learning have been published in recent years to tackle these problems and to support clinicians in identifying and categorizing specific gait patterns into clinically relevant categories~\cite{chau_review_2001-1},~\cite{chau_review_2001-2}. Machine learning methods employed in this context comprise neural networks~\cite{lozano2010human,zeng2016parkinson,vieira2015software}, support vector machines (SVMs)~\cite{wu2007feature,wu2008pca,levinger2009application}, nearest neighbor classifiers~\cite{mezghani2008automatic,alaqtash2011automatic}, and different clustering approaches (hierarchical, k-means, etc.)~\cite{ferrarin2012gait}. The performance of such methods strongly depends on the input data representation \cite{bengio2013representation}.  Frequently used representations in gait analysis comprise discrete kinematic gait parameters (e.g. local minima and maxima of gait signals and time-distance parameters) \cite{giakas1997time,lafuente_design_1998,alaqtash2011automatic}. Additionally, previous research has shown that global signal representations obtained by principal component analysis (PCA)~\cite{soares2016principal,muniz2009application}, kernel-based~PCA~(KPCA) ~\cite{peng_human_2013,xu_fast_2006} and discrete wavelet transformation~(DWT)~\cite{mezghani2008automatic,alaqtash2011automatic} 
are suitable for subsequent classification~\cite{soares2016principal,mezghani2008automatic}.

%
%

Typical use cases for automatic gait analysis described in the literature show a moderate to high accuracy in distinguishing between different pathologies or patient groups~\cite{alaqtash2011automatic,muniz2009application,wu2007feature,levinger2009application,wu2008pca,soares2016principal,lozano2010human}. However, most of the existing literature investigated rather simple cases such as the differentiation between the affected/non-affected limb in hemiplegic patients~\cite{lemoyne2014implementation}, and the distinction of healthy gait from people with neurological disorders~\cite{alaqtash2011automatic,zeng2016parkinson}, transfemoral amputation~\cite{soares2016principal}, and lower limb fractures~\cite{muniz2009application,lozano2010human}. 
A more complex study is presented in~\cite{williams2015classification}, where several disorders associated with traumatic brain injuries are classified.
The majority of published articles employed kinematic and kinetic data derived from three-dimensional gait analysis (3DGA), which provide a vast amount of kinematic and kinetic information for multiple joints. Drawbacks of such 3DGA measurement systems are the relatively time-consuming data recording, the need for highly trained staff as well as high acquisition and maintenance costs. Therefore, 
such analysis tools are often not suitable for daily use in clinical practice.  

To manage the high patient throughput in rehabilitation centers, a frequently used approach is to combine simple visual inspection or 2D video recordings with the quantification of ground reaction forces (GRF) by force platforms, as changes in the morphology of the GRF waveforms reflect pathological gait~\cite{muniz2009application,alaqtash2011automatic}. 
One major drawback of this approach is the loss of clinically relevant and quantifiable information (e.g. gait kinematics), causing a potential decrease in classification accuracy\cite{schollhorn_identification_2002}. However, such simple approaches are common in clinical practice as they overcome the before-mentioned limitations of 3DGA. 
To date, few attempts have been published that use only  
GRF data for automated gait pattern classification \cite{goh_multilayer_2014,soares2016principal}.
Most of these gait classification approaches show promising results. However, the majority of previous works employed relatively small datasets. Alaqtash et al.~\cite{alaqtash2011automatic}, for example, compared the data of 12 healthy adults to those of patients with cerebral palsy and multiple sclerosis (4 patients each), Muniz and Nadal~\cite{muniz2009application} used data from 38 healthy controls and 13 patients with lower limb fracture, and Soares et al.~\cite{soares2016principal} classified GRF data of 20 able-bodied and 12 patients with transfemoral amputation. Such small datasets make it difficult to train robust and reliable classifiers that are applicable in complex real-world scenarios. Furthermore, a majority of studies \cite{muniz2009application,mezghani2008automatic,goh_multilayer_2014} 
relies solely on the vertical ground reaction force for classification purposes, rather than considering all available GRF components, including the center of pressure (COP), for a more conclusive picture of the underlying gait pattern. Previous classification attempts mainly focused on the differentiation between specific diseases 
rather than drawing a distinction between functional gait disorders. 
The work of K\"ohle and Merkl~\cite{kohle2000analyzing, kohle1996identification}, who clustered and classified GRF measurements into deficits of different body regions, represents an exception in this regard.
Their dataset was about half of the size of the one presented in this article and their work also focused on patients walking with a prosthesis.
In this article we define a functional gait disorder as the cause of a gait impairment, which is reflected by the individual gait patterns. These may be associated with a patient's condition after joint replacement surgery, fractures, ligament ruptures, osteoarthritis or related disorders. The classification of functional gait disorders is of particular interest in clinical examinations, as it may play a key role in detecting arthropathies or diseases at an early stage. In addition, such a classification may also indicate secondary disorders that otherwise might be easily overlooked during clinical examination.


The aim of this article is to present a detailed investigation of the automated classification of several functional gait disorders solely based on GRF data. 
The presented approach builds upon the aforementioned studies, e.g. \cite{goh_multilayer_2014, muniz2009application, soares2016principal}, investigates the suitability of frequently used state-of-the-art GRF parameterization techniques for gait classification and analyzes their discriminative power. In the experiments we evaluate the individual representations on a large-scale and real-world dataset for different classification tasks. 
This paper therefore presents a first performance baseline for the automatic classification of different gait disorders in a real-world setting. 

%% file: 04_methods.tex

\subsection{Patients and dataset}


The presented retrospective study was approved by the local Ethics Committee of Lower Austria (GS1-EK-4/299-2014). 
The anonymized data used in this study are part of an existing clinical gait database maintained by a rehabilitation center of the Austrian Workers' Compensation Board (AUVA). The AUVA is the social insurance for occupational risks for more than 3.3 million employees and 1.4 million pupils and students in Austria.
The utilized database comprises GRF measurements from 279 patients with gait disorders (GD) and data from 161 healthy controls (N), both of various physical composition and gender (see Table \ref{tab:data set} for details on the dataset). Patients were manually classified into four classes - calcaneus ``C" (n~=~82), ankle ``A" (n~=~62), knee ``K" (n~=~69), and hip ``H" (n~=~66) - by a physical therapist, based on the available medical diagnosis of each patient. Thus, GD refers to C~$\cup$~A~$\cup$~K~$\cup$~H. The individual GD classes include patients after joint replacement surgery, fractures, ligament ruptures, and related disorders associated with the above-mentioned anatomical areas. The most common injuries present in the hip class are fractures of the pelvis and thigh as well as luxation of the hip joint, coxarthrosis, and total hip replacement. The knee class comprises patients after patella, femur or tibia fractures, ruptures of the cruciate or collateral ligaments or the meniscus and total knee replacements. The ankle class includes patients after fractures of the calcaneus, malleoli, talus, tibia or lower leg, and ruptures of ligaments or the achilles tendon. The calcaneus class comprises patients after calcaneus fractures or ankle fusion surgery. All of the above-mentioned injuries may occur individually or in combination within each class.

Each patient performed one or several measurement sessions. In each session, eight recordings for two consecutive steps were performed. Each bilateral recording is referred to as one trial in this paper. Thus, the utilized dataset contains 1,187 sessions comprising 9,496 individual trials (see Table \ref{tab:data set} for details).
\begin{table*}[ht!]
\centering
\caption{Details of the dataset and classes}
\label{tab:data set}
\begin{tabular}{lcccccc}
\hline
Class & Amount & \begin{tabular}[c]{@{}c@{}}Age (yrs.)\\ Mean $\pm$ SD\end{tabular} & \begin{tabular}[c]{@{}c@{}}Body Mass (kg)\\ Mean $\pm$ SD\end{tabular} & \begin{tabular}[c]{@{}c@{}}Sex\\ (m/f)\end{tabular} & Num. Sessions & Num. Trials \\
\hline
\rowcolor{gray!15}
Healthy Control (N) & 161 & 32.4 $\pm$ 13.6 & 74.1 $\pm$ 16.2 & 84/77 & 161 & 1,288 \\
Calcaneus (C) & 82 & 42.4 $\pm$ 9.9 & 84.5 $\pm$ 12.1 & 74/8 & 320 & 2,560 \\
\rowcolor{gray!15}
Ankle (A) & 62 & 40.0 $\pm$ 11.5 & 88.3 $\pm$ 16.9 & 56/6 & 259 & 2,072 \\
Knee (K)& 69 & 41.5 $\pm$ 11.4 & 83.7 $\pm$ 19.6 & 44/25 & 258 & 2,064 \\
\rowcolor{gray!15}
Hip (H) & 66 & 43.6 $\pm$ 14.7 & 81.6 $\pm$ 18.3 & 53/13 & 189 & 1,512 \\ \hline
\textbf{SUM} & \textbf{440} & \textbf{38.4} \textbf{$\pm$ 13.3} & \textbf{80.7} \textbf{$\pm$ 17.3} & \textbf{311/129} & \textbf{1,187} & \textbf{9,496} \\ \hline
\end{tabular}
\end{table*}

\subsection{Data recording and preprocessing}
Gait analysis was performed on a 10m walkway with two centrally embedded force plates (Kistler, Type 9281B12). 
The force plates were placed in a consecutive order, allowing a person to walk across by placing one foot on each plate. Both plates were flush with the ground and covered with the same walkway surface material, so that targeting was not an issue.
During a session, participants walked unassisted and without a walking aid at a self-selected walking speed until a minimum of eight valid recordings were available. 
These recordings were defined as valid by a supervisor when the participant walked naturally and there was a clean foot strike on the force plate.
Prior to the gait analysis session, each participant underwent rigorous physical examination by a physician. 

All processing steps and subsequent analyses were performed in Matlab 2016a (The MathWorks Inc., Natick, MA, USA). The three analog GRF signals as well as the two COP signals were converted to digital signals using a sampling rate of 2000Hz and a 12-bit analog-digital converter (DT3010, Data Translation Incorporation, Marlboro, MA, USA) with a signal input range of $\pm$10V. A threshold of 10N was used for step detection and 30N for COP calculation. Raw signals were filtered using a 2nd order low-pass butterworth filter with a cut-off frequency of 20Hz. All gait measurements were temporally aligned so that they all started with the initial contact and ended with toe-off. They were further time-normalized to 100\% stance by re-sampling the data to 1000 points. The processed signals are referred to as waveforms in this article. Amplitude values of the three force components, i.e. vertical (V), medio-lateral (ML), and anterior-posterior (AP), were expressed as a multiple of body weight ($BW$) by dividing the force by the product of body mass times acceleration due to gravity ($g$). The COP waveforms from each trial were normalized by the foot length ($FL$) determined during each session, expressed as a multiple of foot length.

\subsection{Signal representation}
The representations employed in our investigation comprise (1)~discrete GRF parameters ~(DP) in combination with time-distance parameters~(TDP) \cite{lafuente_design_1998, alaqtash2011automatic,giakas1997time}; (2)~PCA-based parameterizations of the entire GRF waveforms \cite{lozano2010human,soares2016principal,wu2008pca} and (3)~a combination of the first two approaches, i.e. PCA applied to 
DPs and TDPs \cite{wu2007feature}. In the following, all three approaches are described in detail.

\begin{figure}
  \centering
	\includegraphics[width=0.90\linewidth]{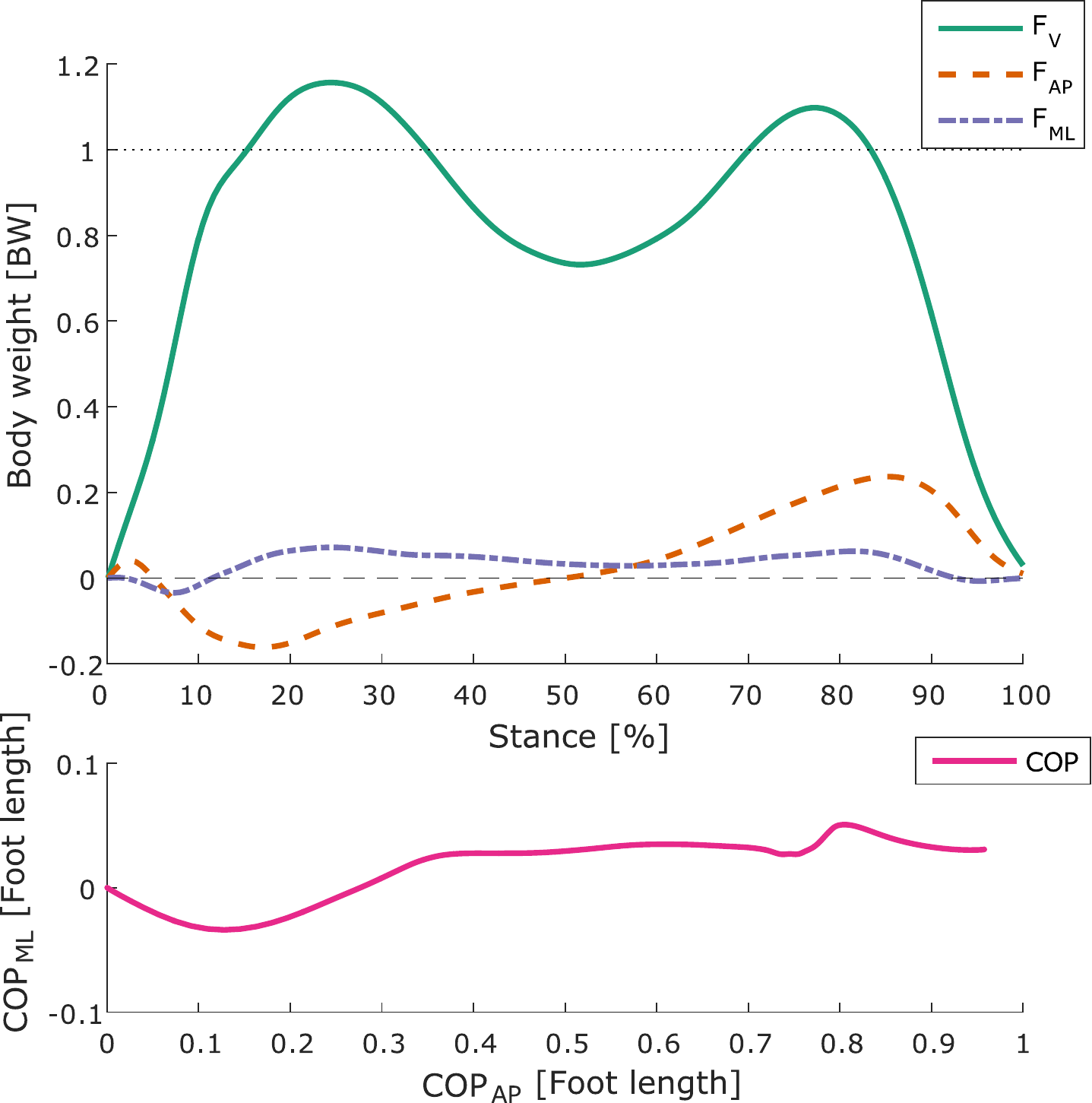}
    \caption{(top) The characteristic shape of the three components of the GRF: the vertical force ($F_{V}$), the anterior-posterior shear ($F_{AP}$), and the medio-lateral shear ($F_{ML}$). (bottom) The corresponding COP path for one step. Note that x and y axes are scaled slightly differently for better visualization.}
    \label{img:GRFcomp}
\end{figure}

DPs were calculated for the affected limb and extracted from all three force components, $F_{V}(t)$ (vertical), $F_{AP}(t)$ (anterior-posterior), and $F_{ML}(t)$ (medio-lateral), as well as from the COP displacement in the anterior-posterior (walking) direction $COP_{AP}(t)$ and in the medio-lateral direction $COP_{ML}(t)$. An example of the GRF and corresponding COP waveforms is presented in Figure \ref{img:GRFcomp}. Furthermore, a more detailed visualization of the mean GRF waveforms over each class is illustrated in Figure S1 (supplementary material). DPs include a set of predefined (most prominent) local minima and maxima of the waveforms, which were extracted by peak detection in a fully automatic way from each trial. Furthermore, impulses were calculated over different segments of the waveform by multiplying the average force (in $N$) by the time this force is active. To account for differences in body mass between participants \cite{moisio_normalization_2003}, all impulses were  divided by the product of body mass times acceleration due to gravity~($g$) and then multiplied by 100 ($\%BW\cdot$$s$). TDPs such as cadence~($CAD$), double support time~($DS$), gait velocity~($GV$), step length~($STEPLEN$), and stance time~($ST$) were calculated from two consecutive steps (affected and unaffected limb) and averaged over the eight valid trials. Table \ref{tab:GRFparams} lists all 52 extracted parameters.

\begin{table*}[ht!]
\centering
\caption{Discrete and time-distance parameters, description, type of normalization and physical unit.}
\label{tab:GRFparams}
\resizebox{.80\textwidth}{!}{%
\begin{tabular}{llll} 
\hline
Abbreviation		&	Description                                            								&	Normalization 		& 	Unit 	\\ \hline
\rowcolor{gray!15}
$ST$				&	Stance time is the duration of the stance phase	of one foot										&	-					&	$s$\\
$F_{V1}$		& 	Maximum value of $F_{V}$ within the breaking phase of stance													&	Body weight			&	$BW$\\
\rowcolor{gray!15}
$T_{V1}$     	&  	Time of $F_{V1}$																				&	Stance time			&	\%ST\\
$F_{V2}$   		&   Minimum value of $F_{V}$ between $T_{V1}$ and $T_{V3}$											&	Body weight			&	$BW$\\
\rowcolor{gray!15}
$T_{V2}$    	&   Time of $F_{V2}$																				&	Stance time			&	\%ST\\
$F_{V3}$  		&	Maximum value of $F_{V}$) within the propulsive phase of stance													&	Body weight			&	$BW$\\
\rowcolor{gray!15}
$T_{V3}$ 		&	Time of $F_{V3}$																				&	Stance time			&	\%ST\\
$F_{AP1}$		&	Maximum value of $F_{AP}$ between initial contact and $T_{AP2}$								&	Body weight			&	$BW$\\
\rowcolor{gray!15}
$T_{AP1}$		&	Time of $F_{AP1}$																				&	Stance time			&	\%ST\\ 
$F_{AP2}$		&	Minimum value of $F_{AP}$) within the breaking phase of stance 		 							&	Body weight			&	$BW$\\ 
\rowcolor{gray!15}
$T_{AP2}$		&  	Time of $F_{AP2}$              																	&	Stance time			&   \%ST\\
$F_{AP3}$		&	Maximum value of $F_{AP}$) within the propulsive phase of stance									&	Body weight    		&   $BW$\\
\rowcolor{gray!15}
$T_{AP3}$		&	Time of $F_{AP3}$                     															& 	Stance time			&   \%ST\\
$F_{ML1}$		&	Minimum value of $F_{ML}$ within the breaking phase of stance				           			&	Body weight			&	$BW$\\
\rowcolor{gray!15}
$T_{ML1}$		&	Time of $F_{ML1}$												                      			&   Stance time			&	\%ST\\
$F_{ML2}$		&	Maximum value of $F_{ML}$ within the breaking phase of stance									&	Body weight     	&	$BW$\\
\rowcolor{gray!15}
$T_{ML2}$		&	Time of $F_{ML2}$          												            			&   Stance time			&	\%ST\\
$F_{ML3}$		&	Maximum value of $F_{ML}$ within the propulsive phase of stance									&	Body weight			&	$BW$\\
\rowcolor{gray!15}
$T_{ML3}$		&	Time of $F_{ML3}$        												              			&	Stance time			&	\%ST\\
$F_{VAVG}$ 		&	Mean value of $F_{V}$																			&	Body weight   		&	$BW$\\
\rowcolor{gray!15}
$F_{APAVG}$ 	&	Mean value of $F_{AP}$																			&	Body weight			&	$BW$\\
$F_{MLAVG}$ 	&	Mean value of $F_{ML}$																			&	Body weight			&	$BW$\\
\rowcolor{gray!15}
$IF_{V}$ 		&	Impulse of $F_{V}$ during stance				 										&   Body weight            		&   $\%BW\cdot$$s$\\	
$IF_{AP}$    	&	Impulse of $F_{AP}$ during stance													&   Body weight            		&   $\%BW\cdot$$s$\\
\rowcolor{gray!15}
$IF_{ML}$   	&	Impulse of $F_{ML}$ during stance													&   Body weight            		&   $\%BW\cdot$$s$\\
$IF_{V1}$ 		&	Impulse of $F_{V}$ between initial contact and $T_{V1}$ 	    						&   Body weight            		&   $\%BW\cdot$$s$\\
\rowcolor{gray!15}
$IF_{V2}$ 		&	Impulse of $F_{V}$ between initial contact and $T_{V2}$ 	  							&   Body weight            		&  	$\%BW\cdot$$s$\\	
$IF_{V3}$  		&	Impulse of $F_{V}$ between initial contact and $T_{V3}$ 	  							&   Body weight            		&  	$\%BW\cdot$$s$\\
\rowcolor{gray!15}
$IF_{APDEC}$	&	Impulse of $F_{AP}$ during the breaking phase										&   Body weight            		&  	$\%BW\cdot$$s$\\     
$IF_{APACC}$	&	Impulse of $F_{AP}$ during the propulsive phase 										&   Body weight            		&  	$\%BW\cdot$$s$\\   
\rowcolor{gray!15}
$IF_{LAT}$   	&	Impulse of the lateral component of $F_{ML}$											&   Body weight            		&  	$\%BW\cdot$$s$\\
$IF_{MED}$   	&	Impulse of the medial component of $F_{ML}$    	         							&   Body weight            		&	$\%BW\cdot$$s$\\
\rowcolor{gray!15}
$COPANG$          &	COP angle is the horizontal angle between the COP linear regression line and the x-axes ($\neq$ foot rotation)					&   -            		&   deg \\
$COPDEV$          &	COP deviation is the root mean square error of the COP linear regression 						&   Foot length            		&   $FL$ \\
\rowcolor{gray!15}
$COP_{AP}$         	&	COP range is the range in the anterior-posterior direction during stance phase                    														&   Foot length            		&	$FL$ \\
$COPV$   			&	COP velocity is calculated as the ratio of foot length and stance time 							      											&   Foot length            		& 	$FL/s$\\
\rowcolor{gray!15}
$COP_{ML}$          	&	COP range is the range in the medio-lateral direction during stance phase                 							&   Foot length            		& 	$FL$\\
$DECT$      		&	Deceleration time (breaking phase) is the duration of $F_{AP}$ being negative                   &   -            		&   $s$\\
\rowcolor{gray!15}
$ACCT$       		&	Acceleration time (propulsive phase) is the duration of $F_{AP}$ being positive        			&   -            		&   $s$\\

$LR0080$   		&	Loading rate represented as the slope of $F_{V}$ from the initial contact to 80\% of $F_{V1}$	&   Body weight            		& 	$N/s$\\
\rowcolor{gray!15}
$LR2080$  		&	Loading rate represented as the slope of $F_{V}$ from 20\% to 80\% of $F_{V1}$											&   Body weight            		&   $N/s$\\

$UR8000$     		&	Unloading rate represented as the slope of $F_{V}$ from 80\% of $F_{V3}$ to the toe-off         &   Body weight            		&   $N/s$\\

\rowcolor{gray!15}
$UR8020$     		&	Unloading rate represented as the slope of $F_{V}$ from 80\% to 20\% of $F_{V3}$                 						&   Body weight            		&   $N/s$\\

$DS$     			&	Double support time during one stride														&   -            		&   $s$\\
\rowcolor{gray!15}
$STEPLEN$			&	Step length is the distance of the COP position from initial contact to following contralateral initial contact &   -      				& 	$m$\\

$STEPV$ 			&	Step velocity is calculated as the ratio of step length and step time								&	-					& 	$km/h$\\
\rowcolor{gray!15}
$STRIDET$         	&	Stride time is the duration from initial contact to initial contact of the ipsilateral foot 		&	-					& 	$s$\\
$BF$        		&	Basic frequency is the mean number of strides per second (1/$STRIDET$)                          	&	-					& 	$Hz$\\
\rowcolor{gray!15}
$CAD$          	&	Cadence is the number of steps per minute														&   -            		&	$1/min$\\
$STEPWD$    		& 	Step width is the medio-lateral distance of the mean COP between both feet						&   -           	 		&	$m$	\\
\rowcolor{gray!15}
$STRLEN$  		&	Stride length is the distance of the COP position from initial contact to following ipsilateral initial contact 		&  -             		&	$m$	\\
$GV$				&	Gait velocity is calculated as the mean step velocity of both feet									&	-					&	$km/h$\\ \hline

\multicolumn{4}{l} {Body weight (BW): product of body mass and acceleration due to gravity;} \\
\multicolumn{4}{l} {\%ST: percentage of stance time; \%BW: percentage of body weight; FL: multiple of foot length.}

\end{tabular}
}
\end{table*}

In contrast to the GRF parameters (DPs and TDPs), the PCA takes the entire waveforms\footnote{For the purpose of the present study, every third sample was used in order to reduce redundancy in the data, thereby improving the robustness of the decomposition.} of the affected limb into account and 
provides a holistic representation of the data. 
Complementary information to the parameters is thus captured. The main goal of PCA is to reduce the dimensionality of a dataset by transforming the data into a set of uncorrelated variables, i.e. the principal components (PCs) \cite{jolliffe2002principal}. Each PC points in (and thus explains) one orthogonal direction of variance in the data. 
%
%
The main intention is to obtain a lower-dimensional representation of our time- and weight-normalized waveforms similar to \cite{lozano2010human,soares2016principal,wu2008pca} by projecting the data onto those PCs which explain most variance in the data. This dimensionality reduction fosters subsequent machine learning\cite{chau_review_2001-2}. 
We performed PCA on each of the five signals separately and concatenated the resulting PCs to obtain a feature vector for classification. This approach proved to be superior to other PCA-based representations in a preliminary study~\cite{slijepcevic2017ground}.
The final dimensionality of the obtained representations is specified by the amount of variance preserved in a particular projection, i.e. 98\%, 95\%, and 90\%. An exemplary visualization of the different PCA representations is presented in Figure S2 (supplementary material). 
A preliminary evaluation indicated that preserving 98\% of the variance results in a good trade-off between data reduction and classification performance. Thus, all results presented in the following are based on the approach in which 98\% of the variance is preserved (the number of resulting PCs is waveform-specific and ranges from four to twelve. For all five signals there are 39 PCs in total). 

As a third representation, PCA was applied to the previously extracted DPs and TDPs (a vector comprising of 52 parameters), similarly to Wu et al. \cite{wu2007feature}. This approach combines both methodologies and aims at extracting the most important information from the (possibly redundant) parameters. 


\subsection{Statistical analysis}

Our first aim was to investigate the suitability of different parameterization techniques for subsequent gait classification. For this purpose we analyzed the variance and discriminative power of each DP and TDP across the different classes by descriptive statistics in a first step. We calculated the median, interquartile-range (IQR) and range of each parameter within each class and visualized them as boxplots. This enabled us to visually inspect variances and distributions in and across the classes, thereby allowing a first estimation of the discriminative power of each parameter.

In a second step, we investigated the discriminative power of the parameters and the global PCA-based representations by linear discriminant analysis~(LDA). A natural measure to describe the separation of two distributions (classes) is the Fisher criterion, which represents the core of LDA \cite{duda2012pattern}. We applied (multi-class) LDA to assess the discriminative power of individual parameters for two (or more) classes. The advantage of this approach is that the discriminative power of a parameter (even across multiple classes) can be expressed by one scalar value that directly reflects the statistical  properties of the input data. 
Hence, there is no need to apply additional modelling and data transformations (which may influence results) prior to LDA, which would be necessary for other methods such as SVM. Furthermore, this approach can easily be extended to estimate the discriminative power of a combination of several parameters by multi-dimensional LDA (e.g. in case of PCA-based representations). We computed the accuracy of LDA and reported the divergence from a random baseline \cite{Vries2012document} to quantify to which degree an input parameter or input representation is able to separate the underlying classes. The random baseline was estimated by the zero rule (always choosing 
the most frequent class in the dataset). Thus, in the case of five classes where the largest class contains 30\% of the data the random baseline equals 
30\%.

\subsection{Classification}

We applied two classification tasks to the dataset by using SVMs as classifiers: (1) (binary) classification between normal gait and all gait disorders ($N/GD$) and (2) (multi-class) classification between N and each of the four GD classes  ($N/C/A/K/H$). In the first task, the class priors are imbalanced, i.e. there are many more observations in the combined GD class than in the normal class (see Table \ref{tab:data set}). The second task separates each type of disorder from each other and from the normal class. 

For the classification experiments the dataset was split into a training (65\%) and a test set (35\%), thereby mutually disjoining the groups of patients in both sets. 
The training dataset in combination with a k-fold cross-validation approach served to train the classifiers and to optimize their parameters (model selection), whereas the test dataset was used to evaluate the generalization ability of the trained models (and was not considered during model selection and hyper-parameter optimization).
The calculated DPs and TDPs as well as the PCA-based representations served as input to classification. The parameters (DPs and TDPs) were normalized (each independently) in a twofold way, by min-max normalization and z-standardization, in order to determine the more suitable approach. The PCA representations were z-standardized. We employed SVMs for the classification with linear and radial basis function (RBF) kernels, provided by the LIBSVM library \cite{ChangLIBSVM2011}. For hyper-parameter selection we applied a grid search over the regularization parameter $C \in [2^{-5}, 2^{15}]$ for the linear SVM and over $C \in [2^{-1}, 2^{15}]$ and the kernel hyper-parameter $\gamma \in [2^{-15}, 2^{5}]$ for the RBF SVM. During the grid search, a 5-fold cross-validation was performed on the training dataset. Finally, an SVM with the best parameters estimated during model selection was trained on the entire training set and evaluated on the test set. Additionally a k-nearest neighbor (k-NN) classifier and a multi-layer perceptron (MLP) were employed as a reference to compare their results to the performance of the SVM. Grid search was performed over various values of k for the k-NN. For the MLPs different numbers and sizes of hidden layers were employed.
As a performance measure we use the classification accuracy, which is the percentage of correct classifications among all classes and input samples. Since in different experiments the random baseline varies, the absolute values of accuracy are of limited expressiveness. To enable a fair comparison, we employ the \emph{divergence from a random baseline} approach  \cite{Vries2012document} and thus provide for each experiment the difference between the random baseline and the absolute classification accuracy. 

%% file: 05_results.tex

\IEEEPARstart{T}{his} section presents and discusses the results of the statistical analysis and the classification experiments.




\subsection{Statistical analysis}

The statistical analysis aimed at assessing the suitability of the individual GRF parameters (DPs and TDPs) for distinguishing different classes of gait disorders. In order to be considered 
a "good" parameter, intra-class variation should be low (e.g. small IQR inside a given class), while the inter-class variation should be high (e.g. significantly different means or medians between the samples of different classes) \cite{lafuente_design_1998}.
%
%

\begin{figure}[h!]
  \centering
	\includegraphics[width=0.95\linewidth]{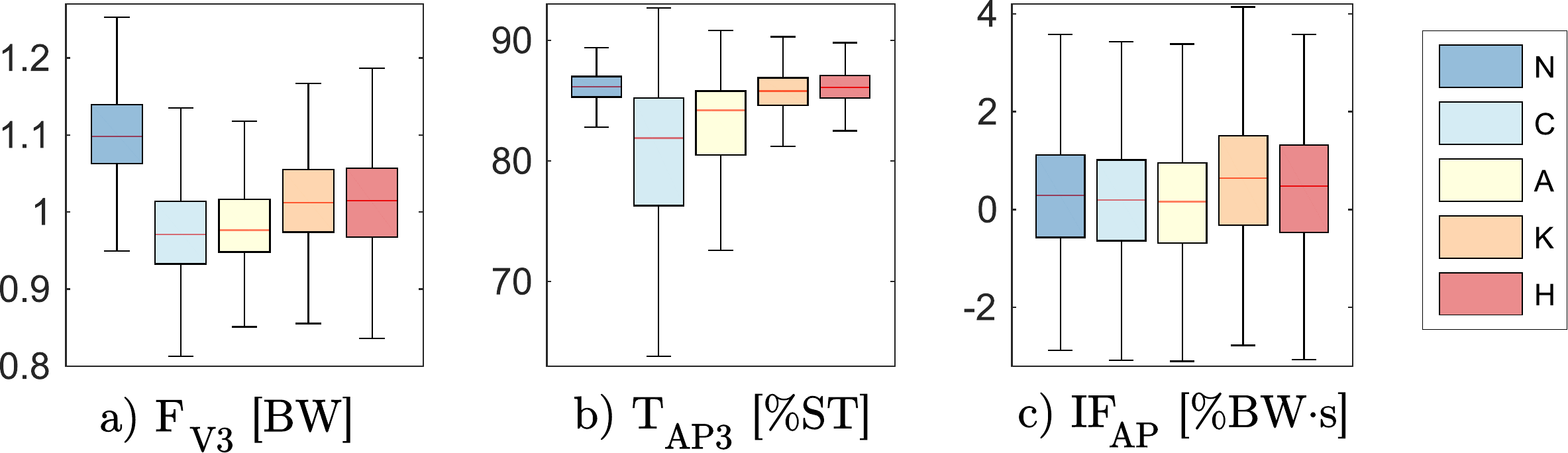}
    \caption{Example of boxplots for three parameters. Each boxplot shows the median and the IQR (box) for each class (outliers were removed for better visualization). Box-whiskers correspond to 1.5 of the box-length, thus show approximately $\pm$~2.7 standard deviations. The overlap of distributions between the classes gives an impression of the parameters' discriminative power.}
    \label{img:variability}
\end{figure}

The visual inspection of the boxplots for each parameter enables a first assessment of the intra- and inter-class variation and thereby gives an impression of the parameters' potential to differentiate between different classes. Figure~\ref{img:variability} shows boxplots for selected parameters. A presentation of boxplots for all 52 investigated parameters for all classes is provided in Figure S3 (supplementary material). 
Parameters such as $F_{V3}$ (see Figure~\ref{img:variability}(a)) show a clear difference in the median and the IQR between the healthy controls and all four GD classes. This indicates a high potential to discriminate between normal gait and arbitrary gait disorders. 
However, the overlap of the distributions within the GD classes indicates a low potential to discriminate between them. Other parameters such as $T_{AP3}$ (see Figure~\ref{img:variability}(b)) vary strongly in the IQR and the median across the classes. While the IQR is high for calcaneus and ankle, the classes hip, knee, and the normal controls exhibit a very similar distribution. Thus, such a parameter has solely limited potential to separate normal gait from general gait disorders. There may be, however, a certain potential to separate individual classes (in this case calcaneus) from other classes. 
Other parameters may lack in discriminative power. An example is $IF_{AP}$ (see Figure~\ref{img:variability}(c)), which shows a similar median and overlapping distributions with a similar IQR across all classes.
Several parameters are discriminative for particular classes or a group of classes. However, none of the observed parameters discriminates well between all classes. Therefore, the combination of several parameters for the distinction between classes seems advisable. 
These assumptions are further corroborated by the LDA results.

\begin{figure*}[ht!]
  \centering
    \includegraphics[width=\textwidth]{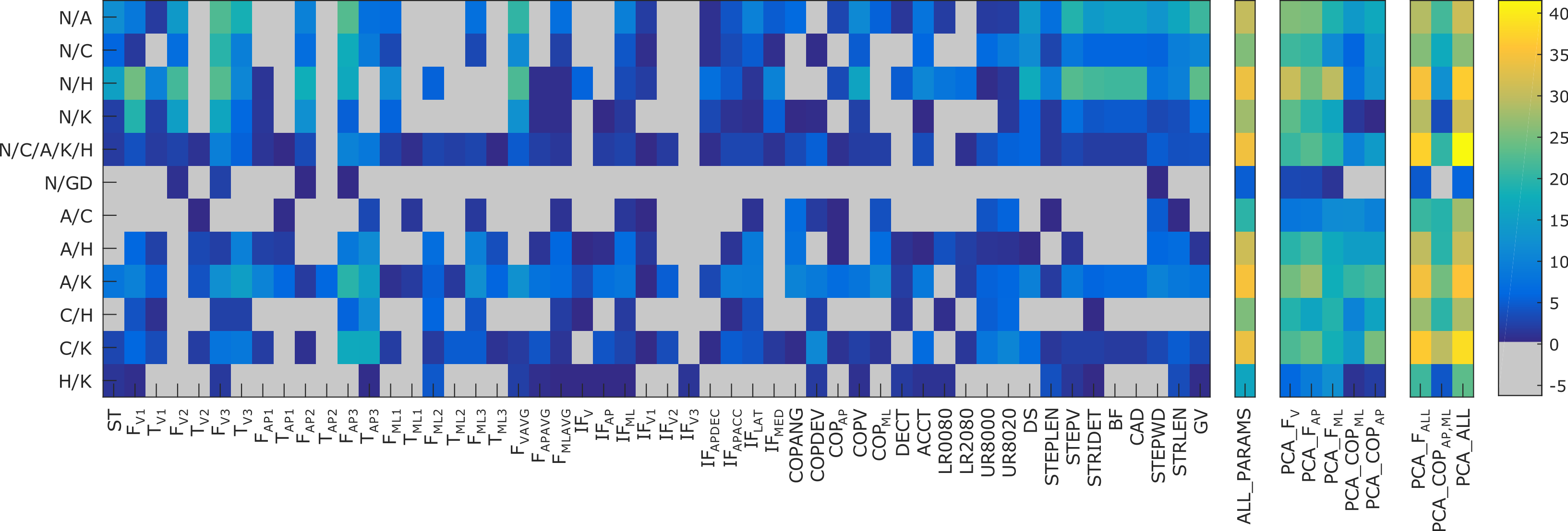}
    \caption{Discriminativity scores obtained by LDA for different selections of classes (rows). The figure is divided into four blocks. Each column represents a different input parameter or higher-dimensional input representation. Best viewed in color.}
    \label{img:discriminativity}
\end{figure*}

LDA was applied to the individual parameters and their combination, as well as to the higher-dimensional PCA-based representations. This analysis aimed at quantifying the discriminative power of the investigated representations and thereby evaluating their suitability for automated classification. Figure \ref{img:discriminativity} illustrates discriminativity scores obtained by LDA in terms of deviation from the random baseline (\textit{zero~rule}). In detail, results for different combinations of classes (rows) are illustrated: rows~1-4 provide results for the discrimination of normal gait vs. ankle, calcaneus, hip or knee (each class separately). Row~5 shows how well all 5 classes can be differentiated from each other. Row~6 illustrates how well normal gait can be differentiated from all types of gait disorders. Rows~7-12 show how all possible pairs of gait disorder classes can be differentiated from each other. 
%
%
Positive discriminativity scores are represented by a color scale from blue (corresponding to low values) to yellow (representing high values), whereas negative values are colored in gray. Positive values mean that the random baseline is exceeded and that the respective input parameter or input representation exhibits a certain discriminative power (the higher the value the better). Negative values indicate the absence of discriminative power, i.e. the random baseline is not reached. It has to be noted that, since the different class partitions represented by the rows of Figure~\ref{img:discriminativity} have different random baselines, the values across rows cannot be compared directly. Comparisons are solely valid along the rows. In general, however, columns including a larger number of high values indicate parameters or representations with a higher discriminative power. Similarly, rows with higher values represent tasks that are easier to solve than others.

The leftmost part of Figure~\ref{img:discriminativity} illustrates the discriminativity scores for the individual parameters. Several parameters achieve high scores for individual classes or combinations of classes, e.g. $F_{AP3}$, $F_{V3}$, $F_{VAVG}$, $F_{V1}$, $T_{V3}$, $T_{AP3}$, $GV$, $STEPV$, $DS$, $STRLEN$, $F_{V2}$, $STEPWD$, $CAD$, $BF$, and $STRIDET$. No parameter, however, performs well across all tasks. This indicates that individual parameters are quite limited in expressiveness. The second part ($ALL\_PARAMS$) of Figure~\ref{img:discriminativity} illustrates the results from the combination of all parameters. The combination yields much better discrimination across all rows of Figure~\ref{img:discriminativity}. This demonstrates that the individual parameters contain complementary information and attain synergies when they are combined. The third part of Figure~\ref{img:discriminativity} visualizes the results of the PCA-based representations of the five input signals $F_V$, $F_{AP}$, $F_{ML}$, $COP_{AP}$, and $COP_{ML}$. The three GRF components achieved higher scores compared to the COP signals. 
The rightmost part of Figure~\ref{img:discriminativity} shows the discriminativity scores for combined PCA representations, i.e all three GRF components combined ($PCA\_F_{ALL}$), both COP signals combined ($PCA\_COP_{AP,ML}$), and all five components combined ($PCA\_{ALL}$). In general, the combination of components improved the results, which indicates 
that the individual GRF components are complementary to each other. The addition of the COP further improved the discriminative power. Thus, adding COP to a classification may contribute positively to the results. The 
representations ($PCA\_{ALL}$ and $ALL\_PARAMS$) are 
combined able to contribute to all evaluated tasks (rows) of Figure~\ref{img:discriminativity}.


The evaluated representations are more suitable for differentiating between the healthy control group and a functional gait disorder (rows 1-4) than between two functional gait disorders (rows 7-12). Regarding the task $N/GD$, solely a few parameters are able to exceed the random baseline. 
This is due to the fact that the combined set of all gait disorders contains much more samples than the class of healthy controls (i.e. 279 vs. 161 samples). This yields a random baseline around 87.1\% which is more difficult to exceed than random baselines in other tasks.




\subsection{Classification}
\label{subsec:classification}

The results of the classification experiments, which were performed on data from the test set, are summarized in Table \ref{tab:results}. The test set was not presented to the classifier during the training phase and the selection of its parameters. Results are provided for the two classification tasks ($N/C/A/K/H$ and $N/GD$) and for three different parameterizations. 
The results of the additional experiments with other classifiers such as the multi-layer perceptron (MLP) and the k-nearest neighbors algorithm (k-NN) were all outperformed by the SVM results, which confirms also the results of Janssen, Sch{\"o}llhorn et al. \cite{janssen2011diagnosing}. Therefore, and due to the limited space available, these results will not be discussed in detail.

The first evaluated parameterization comprises of 52 GRF parameters (DPs and TDPs) that are extracted from all five GRF input signals. Due to the strong variation in the parameters' value ranges, a suitable normalization of the data is essential. We evaluated min-max normalization as well as z-standardization. The use of z-standardization resulted in a slightly higher deviation from the RB 
for both tasks (except for the RBF SVM in task $N/C/A/K/H$) compared to min-max normalization. Furthermore, the RBF SVM failed to exceed the random baseline for both methods in the task $N/GD$.

The second parameterization was obtained by PCA of the raw GRF waveforms. PCAs obtained solely from the three force components clearly outperform the GRF parameters (DPs and TDPs). By adding the COP measurements the results were further improved for both tasks. Normalization of the PCA-based representations is crucial as performance otherwise drops significantly.

The third parameterization applied PCA on the z-standardized and min-max normalized DPs and TDPs. The dimensionality reduction resulted in a 28-dimensional vector which was also z-standardized prior to classification. In this case, results for both normalizations (last two rows of Table \ref{tab:results}) were improved for the task $N/GD$ compared to the representation with the original GRF parameters (first two rows of Table \ref{tab:results}).
However, this is not the case for 
task $N/C/A/K/H$, where the deviation from the RB 
slightly decreased. 

In summary, the best performance 
(marked in bold in Table \ref{tab:results}) was achieved by applying PCA to all five GRF signals. The linear SVM achieved the highest deviation from the RB (22.5\%) for 
task $N/C/A/K/H$ as well as for 
task $N/GD$ (3.7\%).
Alternative classifiers which were also evaluated yielded a lower deviation from the RB: MLP~21.0\% and k-NN~13.4\% for 
task $N/C/A/K/H$ and MLP~2.6\% and k-NN~2.2\% for 
task $N/GD$. In terms of accuracy and deviation from the RB, the linear SVM performed better in all experiments. The RBF SVM has an advantage solely in terms of runtime.

The main 
reason for the great difference in the performance  between the two tasks is the strong class imbalance in task $N/GD$, which makes this task particularly difficult to solve. 
One way of dealing with unbalanced datasets in SVMs is the use of different weights for different classes, thereby emphasizing the importance of the under-represented classes. Therefore, additional class-weighted experiments were performed. Results with different cost functions showed that no further performance increase can be achieved. The uniform cost function seems to work best on the data.

\begin{table*}[ht!]
\centering
\caption{Classification results (\%) of two tasks - $N/C/A/K/H$ and $N/GD$ - and three different parameterization approaches. Note that the random baseline (RB) is stated next to the task name and that the values in the table represent the deviation from the random baseline (RB) and the corresponding absolute accuracy in brackets.}
\label{tab:results}
\begin{tabular}{llc|cccc}
\hline

\multicolumn{1}{l}{\multirow{2}{*}{\textbf{Parameterization}}} & \multicolumn{1}{l}{\multirow{2}{*}{\textbf{Norm.}}} & \multicolumn{1}{c|}{\multirow{2}{*}{\textbf{Dim.}}} & \multicolumn{2}{c}{$N/C/A/K/H$~(RB:~31.8\%)}  & \multicolumn{2}{c}{$N/GD$~(RB:~87.1\%)} \\ \cline{4-7} 

\multicolumn{3}{c|}{} 			& \multicolumn{1}{c}{\textbf{linear SVM}} & \multicolumn{1}{c}{\textbf{RBF SVM}}  & \multicolumn{1}{c}{\textbf{linear SVM}}   & \multicolumn{1}{c}{\textbf{RBF SVM}}     \\ \hline
\rowcolor{gray!15}
GRF Parameters (DPs and TDPs)	&	z-score	&	52 & 15.0 (46.8) 	& 8.8	(40.6) &  2.4 (89.5) & -0.8 (86.3)\\
GRF Parameters (DPs and TDPs)&	min-max	&	52 & 14.3 (46.1)	&	9.5 (41.3)  & 1.6 (88.7)	& -3.8 (83.3)\\
\hline
\rowcolor{gray!15}
PCA on $F_{V}$, $F_{AP}$, $F_{ML}$ & z-score&	30	& 19.8 (51.6)	&	15.4 (47.2) & 2.4 (89.5) & \textbf{2.0 (89.1)}\\
PCA on $F_{V}$, $F_{AP}$, $F_{ML}$, $COP_{AP}$, $COP_{ML}$ & z-score&	39 & \textbf{22.5 (54.3)} &	\textbf{19.4 (51.2)} & \textbf{3.7 (90.8)} & 1.9 (89.0)\\
\hline
\rowcolor{gray!15}
PCA on z-standardized GRF parameters &	z-score&	28 & 13.8 (45.6) &	8.8 (40.6) & 2.6 (89.7) & -0.6 (86.5)\\
PCA on min-max normalized GRF parameters &	z-score&	28 & 13.5 (45.3) &	7.9 (39.7) & 2.8 (89.9) & 0.1 (87.2)\\
\hline
\end{tabular}
\end{table*}


\subsection{Discussion and further aspects} 
We presented a study on the classification of different functional gait disorders, stemming from a wide range of possible impairments, into categories that represent the main affected body region. The motivation for selecting these broad categories is that identifying the region of impairment is essential for clinical practice and may allow to pinpoint impairments already at an early stage. In addition, it could indicate secondary impairments which may easily be overlooked by the physician during clinical examination. The present study represents a first performance baseline for the classification of gait disorders. Results are particularly promising for 
task $N/GD$. However, an absolute classification accuracy of 91\% still lies below an acceptable threshold for clinical practice. For the classification of individual disorder categories, the results indicate that further improvements are necessary. To date, the proposed approach could, however, already serve as a support for clinicians indicating the presence of (additional) arthropathies or diseases. 
In order to reduce the classification complexity, while still providing support for clinicians, similar classes could be merged, i.e. the hip and knee classes into a thigh class and the ankle and calcaneus classes into a shank class. The results of this additional experiment showed a deviation from the RB of 26.8\% (using a linear SVM, RB:~51\%, absolute accuracy:~77.8\%). Compared to the distinction of all five classes ($N/C/A/K/H$), this is a clear increase in accuracy and deviation from the RB.

Different influencing factors, i.e. the imbalance of the class priors, the variability in the number of sessions per person and gender-specific aspects may introduce a bias into the aforementioned analyses. To investigate the effect of these factors on classification performance, we performed additional experiments. To this end, we used the best configuration found so far as a baseline, i.e. PCA on all five signals with a linear SVM (4th parametrization in Table \ref{tab:results}) and applied it to different balanced subsets of our dataset. The results are presented in Table~\ref{tab:results2} and are discussed in the following.  

\begin{table}[ht!]
\centering
\caption{Results (\%) of analyses assessing the influence of several factors on the results of the two tasks - $N/C/A/K/H$ and $N/GD$. The experiments are performed with a PCA on all five signals in combination with a linear SVM. Note that the values represent the deviation from the random baseline (RB) and the corresponding absolute accuracy in brackets.}
\label{tab:results2}
\begin{tabular}{l|cc}
\hline
\textbf{Partitions of the dataset}                                           & $N/C/A/K/H$ & $N/GD$ \\
\hline
\rowcolor{gray!15}
Session are balanced                       &     23.7 (60.2)      &  20.6 (84.1)   \\
Persons are balanced                       &      28.3 (59.5)     &      5.3 (84.7)\\
\rowcolor{gray!15}
Persons \& sessions are balanced &     39.2 (59.2)      &    35.4 (85.4) \\
Male population                                                        &     20.9 (51.3)     &   0.6 (91.4)  \\
\hline
\end{tabular}
\end{table}

For the experiments in Section \ref{subsec:classification} we decided to use all available sessions of patients recorded in the course of their rehabilitation to account for different progression stages of impairments. This, however, may introduce a bias in the experiments as  more trials exist for some patients than for others. To evaluate to which extent the varying number of recorded sessions per patient influences the overall result, we balanced the dataset by selecting only one random session per person. Interestingly, the deviation from the RB improved for task~$N/C/A/K/H$ to 23.7\% (+1.2\%) and for task~$N/GD$ to 20.6\% (+16.9\%), as presented in the first row of Table~\ref{tab:results2}. These results show that intra-patient variability needs to be taken into account and requires additional modeling in a classification approach.

Another factor causing an imbalance in the data are the different class cardinalities, i.e. different numbers of persons per class.
In order to investigate the influence of this imbalance we performed an experiment for both tasks with a dataset containing the same number of participants per class (but keeping all sessions in the dataset).  
For task~$N/C/A/K/H$ the balanced dataset is composed of data from 62 persons from each class (overall 310 persons, 7616 trials).
For task~$N/GD$ the balanced dataset contained data from 160 healthy controls and 160 persons with a deficit (40 from each GD class,  overall 320 persons and 6096 trials).
The deviation from the RB improved for task~$N/C/A/K/H$ to 28.3\% (+5.8\%) and for task~$N/GD$ to 5.3\% (+1.6\%), as shown in the second row of Table~\ref{tab:results2}. Although the results show that balancing the number of patients among classes is beneficial, the results of task~$N/GD$ reveal the still existing imbalance in the dataset (due to the fact that healthy controls have only one session and patients up to several sessions).   


The next question deals with the effect of balancing the number of patients and the number of sessions at the same time. We performed experiments with a completely balanced version of our dataset for each task, containing only one session per person and equal numbers of persons per class. For task~$N/C/A/K/H$ the balanced dataset is composed of data from 62 persons from each class (overall 310 persons, 2480 trials). For task~$N/GD$ the balanced dataset contained data from 160 healthy controls and 160 persons with a deficit (40 from each GD class,  overall 320 persons and 2560 trials). The results of our experiments showed clear performance improvements of +16.7\% in the deviation from the RB compared to the baseline for task $N/C/A/K/H$ and +31\% compared to the baseline for task $N/GD$ (see the third row in Table~\ref{tab:results2}). 

Other biases in the data may be introduced by variations in gender, walking velocity, leg length and other parameters \cite{pierrynowski_enhancing_2001} leading to a variability of GRF parameterizations in the individual disorder classes. Additional normalization of the input data may be necessary to reduce intra-class variation to improve classification accuracy. Several studies have shown that in particular gender causes strong variability in gait signals~\cite{eskofier2012pattern,chiu2007effect}. To assess the influence of gender on our results, an experiment was performed on a reduced dataset containing only data from male participants (note that the number of female participants in our dataset is not sufficient to perform separate experiments). Surprisingly, the results did not improve (see the last row in Table~\ref{tab:results2}). This indicates that for our data, gender has rather little influence on the results, which, however, does not imply that the influence of gender can be neglected a priori. A detailed study on the influence of gender is subject to future investigation.




%


%% file: 06_conclusion.tex
\IEEEPARstart{T}{he} present study aimed at classifying patients with different orthopedic gait impairments at the hip, knee, ankle, and calcaneus from healthy controls using GRF measurements. For this purpose a dataset of 9,496 gait measurements from clinical practice was utilized. In a first step we investigated the suitability of state-of-the-art GRF parameterizations and analyzed their statistical properties and discriminative power among the classes. 
Based on these results, the use of entire GRF waveform parameterizations as input (such as PCA), rather than relying on GRF parameters (DPs and TDPs) seems advisable. Furthermore, the use of GRF force components paired with the respective COP measurements yielded the best results. Our experiments further showed that balancing the dataset significantly improves results (e.g. increasing the deviation from the random baseline by +16.7\% for the classification into healthy controls and all four GD classes and by +31\% for distinguishing between healthy controls and patients).
 
The presented study shows that results heavily depend on the employed GRF representation. Future work will investigate and evaluate adaptively learned signal representations \cite{zhang2013pathological,hannink2017sensor} to obtain more discriminative and expressive parameterizations of GRF measurements. Furthermore, we will focus on establishing a large, open-source, and balanced data set to foster further developments in this area. Our results thereby provide a first performance baseline for the classification of functional gait disorders and can serve as a reference for future improvements.